\pdfoutput=1
\documentclass[11pt,a4paper]{article}
\usepackage[hyperref]{eacl2021}
\usepackage{times}
\usepackage{latexsym}
\usepackage{graphicx}
\usepackage{multirow}

\usepackage{microtype}

\aclfinalcopy 

\title{Towards Emotion Recognition in Hindi-English Code-Mixed Data: A Transformer Based Approach}

\author{Anshul Wadhawan and Akshita Aggarwal\\
  Department of Computer Engineering\\
  Netaji Subhas University of Technology\\
  Dwarka, New Delhi \\
  {\texttt \{anshulw, akshitaa\}.co.16@nsit.net.in} \\}

\date{}

\begin{document}
\maketitle
\begin{abstract}
In the last few years, emotion detection in social-media text has become a popular problem due to its wide ranging application in better understanding the consumers, in psychology, in aiding human interaction with computers, designing smart systems etc. Because of the availability of huge amounts of data from social-media, which is regularly used for expressing sentiments and opinions, this problem has garnered great attention. In this paper, we present a Hinglish dataset labelled for emotion detection. We highlight a deep learning based approach for detecting emotions in Hindi-English code mixed tweets, using bilingual word embeddings derived from FastText and Word2Vec approaches, as well as transformer based models. We experiment with various deep learning models, including CNNs, LSTMs, Bi-directional LSTMs (with and without attention), along with transformers like BERT, RoBERTa, and ALBERT. The transformer based BERT model outperforms all other models giving the best performance with an accuracy of 71.43\%.
\end{abstract}


\section{Introduction}
With the growth of social networking sites like Facebook and Twitter, humans have started communicating online much more than ever before. This leads to the generation of huge amounts of textual data which introduces interesting challenges in the domain of NLP. Automatic detection of various linguistic expressions like irony, hate, sarcasm, aggression etc. is being widely explored. Another problem that has drawn keen interest of NLP researchers is detecting emotions of a human via the texts they have produced. In order to aid human-computer interaction, determining the emotions via texts becomes significant \cite{1}. There are multiple ways of detecting emotions, including but not limited to speech \cite{3}, facial expressions recognition \cite{2} and text-based approaches.

Text-based emotion detection is based on the assumption that when a person is happy, they would use positive words. Likewise, when they are angry, frustrated or upset, negative emotions will be depicted by a certain kind of words carrying negative connotation.
Contrary to popular belief, emotions are not only significant in human creativity, but they also play an instrumental part in making rational decisions. With the rise of artificial intelligence and increased focus on human-computer interaction, smart machines that will communicate naturally and intelligently with humans, need to recognize their emotions effectively. Affective computing has emerged as an exciting field with recent focus on emotion detection \cite{4}.

Most of earlier work has been carried out on a mono-lingual dataset due to easy availability of a large corpus of annotated data \cite{5,6}. However, in multilingual cultures, use of multiple languages while exchanging information on social media is quite common. Studies show that as many as 314.9 million people in India are bilingual\footnote{\url{https://en.wikipedia.org/wiki/Multilingualism_in_India}}. This leads to the issue of code-mixing and code-switching especially while communicating on social media  platforms like Twitter, Facebook and Reddit \cite{7,11}. Code-mixing occurs when lexicons and grammatical features of multiple languages are used in the same sentence \cite{8,9,10}.
The major issue in dealing with code-mixed problems is the absence of sufficiently annotated datasets \cite{12}. 

In this paper, we present our findings on one of the most challenging problems in the domain of Natural Language Processing, ‘emotion detection’. While a lot of work has been carried out for the English language \cite{13}, the domain of Hindi-English code-mixed texts remains relatively new and not much explored. We present an annotated Hindi-English code-mixed dataset of 150k tweets for addressing this issue and for enabling future researchers to contribute to this domain. Our aim in this paper is to compare multiple deep learning models including CNNs, LSTMs, Bi-directional LSTMs (with and without attention) with the aid of bilingual self-trained word embeddings on a code-mixed dataset, along with transformer based models like BERT, RoBERTa, and ALBERT. 

The paper is organized as follows – Section 2 details about the background and related work in this domain. Section 3 enumerates the methodology we used to perform the experiments including data annotation, pre-processing, embeddings and models used. Section 4 lists down the experimental settings to replicate the work done. Section 5 contains details of the results obtained and section 6 consists of conclusions drawn from the results. 

\section{Related Work}

With the huge growth of micro-blogging platforms like Facebook and Twitter, there has been an increased interest in detecting sentiments and emotions in large text corpus \cite{14,15}. In initial work aimed at emotion detection in textual data, experiments have been carried out with text-based emotion classification in fairy tales for kids on the lines of basic emotions \cite{16,17}. In another related work  \cite{18}, the authors work on real-world knowledge bases highlighting human’s natural reactions towards various situations, aimed at identifying emotions at the sentence-level. With the increase of non-native English speakers on social media, sentiment analysis on regional languages and code-mixed data has gained momentum.

A pivotal work of sentiment analysis in Hindi corpus was done, where the authors were successful in extracting sentiment lexons from HindiWordNet and were able to achieve an accuracy of 87\% in the domain of movie \cite{19}. In a detailed analysis of data of English-Hindi bilingual users on Facebook, it was shown that 17.2\% of all posts, which accounted for around one-fourth of the words in their dataset, revealed some form of code-mixing \cite{20}. Sub-word level LSTM architecture for performing sentiment analysis was introduced on Hindi-English code-mixed dataset \cite{21}. Experiments were conducted with supervised learning (SVM) on a Hindi-English code-mixed corpus for emotion detection \cite{22}.

\section{Proposed Methodology}

This section describes the series of steps that constitute the methodology proposed, including detailed descriptions of dataset creation, annotation, preprocessing, embeddings, and the deep learning models.

\subsection{Dataset Creation}
The dataset annotated by paper \cite{22} contains 2866 tweets. This data being insufficient for doing any meaningful work with deep learning due to the issue of overfitting, we created a self-annotated class-balanced dataset using TwitterScraper API\footnote{\url{https://github.com/taspinar/twitterscraper}} with relevant search tags like  \#happy, \#sad, \#angry, \#fear, \#disgust, \#wow along with some commonly used hindi words to obtain Hinglish data. 

\subsection{Dataset Annotation and Analysis}
We scraped around 250k tweets for analysis. After dropping the noisy instances containing unknown characters, we filtered the dataset down to a class balanced corpus of 150k tweets. The tweets were annotated with six standard emotions, including, happiness, sadness, anger, fear, disgust and surprise \cite{17}. The hashtags which were used as searching criteria for scraping the tweets, were used for annotation. All examples which were fetched using hashtags like \#yayy were marked to have a positive happiness label. This process was repeated for all the 6 emotions under consideration. 
\begin{table}[h!]
\centering
 \begin{tabular}{|c|c|} 
 \hline
 \textbf{Emotion} & \textbf{Number of instances} \\ 
 \hline
 Happiness & 25869 \\ 
 \hline
 Sadness & 20931  \\
 \hline
 Anger & 28705 \\
 \hline
 Fear & 18981 \\
 \hline
 Disgust & 35667 \\
 \hline
 Surprise & 18935 \\
 \hline
 Total sentences & 149088 \\ 
 \hline
\end{tabular}
\caption{Tweet count per class}
\label{table:1}
\end{table}
The number of tweets per class is depicted in Table 1. Initially, embeddings were trained on just Hinglish tweets, however, English tweets were added later because of excess of hindi words in Hinglish tweets, causing a lack of English specific words. 
The labelled emotion detection dataset along with the deep learning classification models is made available online\footnote{\url{https://github.com/anshulwadhawan/emotion\_detection}} to promote additional research.
\newline


Examples of some annotated data :\newline

TWEET: Great darshan today at siddhi vinayak along wid aarti!! \#happiness @dollydas261 @vishal71182 @vishalbti ..\newline
\indent TRANSLATION: Had a great experience in Siddhi Vinayak Temple, along with the ceremonies \#happiness\newline
\indent EMOTION: Happy\newline

TWEET : Jindagi me Maut sabse bada loss nahi, sabse bada loss tab hota hai jab Do logo ke jinda rehte hue unke beech aapsi riste toot jaye.\#Sad :(
\newline
\indent TRANSLATION: The biggest loss in life is not death. The biggest loss is banishment of relations between loved ones even when alive.\newline
\indent EMOTION: Sad 

\begin{figure*}[h!]
\begin{center}
    \centering
    \includegraphics[width=12cm]{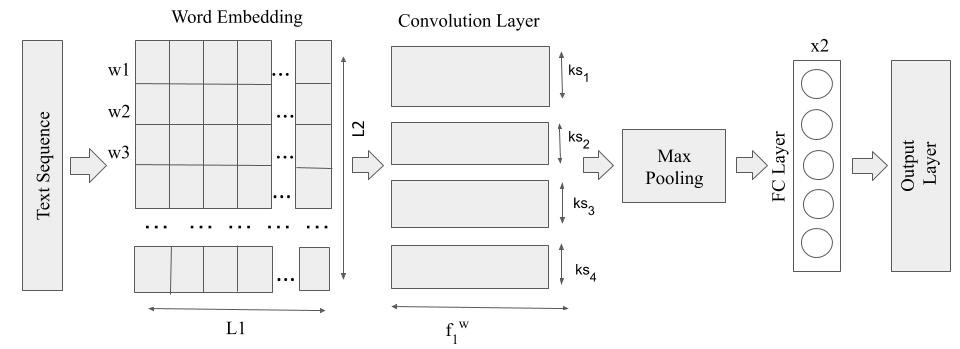}
    \caption{CNN model architecture}
    \label{figure}
\end{center}
\end{figure*}

\subsection{Data Preprocessing}
We preprocessed the scraped data by retaining only Hinglish tweets while removing tweets in pure English and Devanagari. We also removed rare words (words having occurrence of less than 10 in the entire dataset), mentions, ‘\#’ symbols, URLs, punctuations and keywords used for scraping (like happy, sad, etc.) in order to feed our models with cleaner data.
\subsection{Creation of Hindi-English Bi-lingual Word Embeddings}
This being a multi-label text classification problem, it is required that the text be first converted to a form understandable by the various machine learning algorithms. Word embeddings are numerical representation of words. Specifically, word embeddings are vector representations of words that are learned in an unsupervised manner where their relative similarities are directly related to their semantic similarity \cite{23}. Due to unavailability of pre-trained Hindi-English bilingual word embeddings, we created our own embeddings by scrapping 427k Hinglish tweets and 300k English tweets using TwitterScrapper API.
Processing was carried out by removing pure English and pure Devanagari tweets along with rare words, hashtags and mentions for obtaining better training results. We chose 2 types of word embeddings for our problem, each of which was trained on two kinds of datasets, after processing (removing hashtags, user mentions, URLs, punctuations and keywords used for scraping), one which had only Hinglish tweets, the other which had a mix of English and Hinglish tweets. In order to get the right co-relation between the words of the two languages, we experimented with a mixture of Hinglish and English tweets.

{\bf Word2Vec}: In this kind of embedding, words are converted to vector representations where words having common context are placed in vicinity amidst the vector space \cite{24}. Taking a huge corpus of words as input, it generates a vector space with each word being assigned a unique vector value in that space. Since the available Word2Vec embeddings are pre-trained on English datasets only, we trained our embeddings on custom Hindi-English code-mixed dataset, to obtain the desired code-mixed embeddings.

{\bf FastText}: FastText is a modification to the Word2Vec embeddings that was developed by Facebook in 2016 \cite{25}. FastText assumes a word to be composed of character n-grams \cite{26} and hence breaks a given word into various sub-words (Example: light, li, ig, igt, gt) unlike word2vec which feeds individual words to the network. The training session of a FastText model involves learning of weights for not only the whole word, but also for each of the character n-grams. Unlike Word2vec, it can not only approximate rare words but also give representation of words not present in the corpus, as now it is highly possible that some of their n-grams are present in other words. This is particularly useful for messages on social networks where multiple representations are used for similar words (like pyar, pyaar, pyaaar).

\subsection{Deep Learning Models}
We introduce seven deep learning based models for solving the task of emotion detection in textual code-mixed data. The models proposed are CNN, LSTM, Bi-directional LSTM, attention based Bi-directional LSTM and transformer based models like BERT, RoBERTa and ALBERT. We trained FastText and Word2vec word representations on two types of data, one which solely consisted of hinglish text, the other which was a mixture of hinglish and english text. These embeddings were then used to predict the emotion of the tweet by serving as input to all the proposed models except the transformer based models. 

\subsubsection{Convolutional Neural Networks (CNNs)}

CNNs have been proven to be successful for multi class classification problems, where images are provided as inputs \cite{28}. In our case, word embeddings are given as input, from which features are extracted and final classification is performed. The network architecture we employed has been depicted in Fig. 2. Embedding layer serves as the first layer, which is used to transfer the word vector representations of select words in the tweet under consideration, to the model structure. Four convolutional layers in parallel receive the output of the embedding layer, followed by a global max pooling layer, upon which dropout is applied. Three dense fully connected layers follow in which the last layer is responsible for classification. Application of dropout led to better convergence and decreased difference in the training and validation accuracies.

\subsubsection{Recurrent Neural Networks (RNNs)}

The context in which a word is used, determines the meaning of the word, which in-turn may play a significant role in determining the overall sentiment of the sentence. For example, \\
\indent Sentence 1 : There are multiple kinds of human beings in this huge world. \\
\indent Sentence 2 : She is very generous and kind-hearted. \\
The context in which the word 'kind' is used, is different in both the sentences, thus the word carries different meanings in different scenarios. RNNs are helpful in modelling the context of a word, by having unique ways to capture the context of words using the surrounding words. 
\newline



{\bf Long Short-Term Memory (LSTM)}: LSTMs have been shown to capture the relevant context for words \cite{29} as well as address the issue of vanishing gradients \cite{31}. 
\begin{figure}[h!]
\begin{center}
    \centering
    \includegraphics[width=6cm]{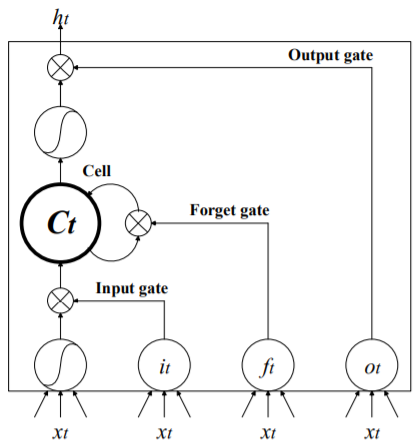}
    \caption{LSTM block structure}
    \label{figure}
\end{center}
\end{figure}
The words which precede a particular word, determine the context of the word. LSTMs inculcate memory cells in the network which serve to record the meaning of words that occurred previously. In order to model this scenario, an LSTM based network is constructed. In our model, an embedding layer, followed by an LSTM layer, further followed by 2 fully connected layers constitute the network. The last layer, consisting of 6 neurons, is responsible for the classification of the tweet’s emotion. 

{\bf Bi-directional LSTM}: Bi-directional LSTMs have been proven successful in capturing the context for text classification tasks \cite{27}. The words that precede as well as follow a particular word, determine the context of the word under consideration. Thus, memory cells must exist in both directions in order to maintain the track of words that surround a particular word. This is achieved by appending 2 LSTM layers to the embedding layer, whose concatenated output (\overrightarrow{h\textsubscript{T}},  \overleftarrow{h\textsubscript{1}}) is flattened and fed to 2 fully connected (FC) layers. The last layer carries out classification, as is done in all other proposed models. 



{\bf Attention based Bi-directional LSTM}: The technique of attention is based on learning the words which contribute the most towards the overall emotion of the sentence, and filtering out the words which contribute the least, i.e. noise.
Attention based BiLSTM differs in the manner of concatenation of states, which is fed to the fully connected (FC) layers. Apart from using concatenated \{\overrightarrow{h\textsubscript{T}}, \overleftarrow{h\textsubscript{1}}\} (\overrightarrow{h\textsubscript{T}} denoting forward directed final hidden state representation, \overleftarrow{h\textsubscript{1}} denoting backward directed first hidden state representation) as inputs to the fully connected layers, attention based BiLSTMs also take into consideration the weighted summation of all time steps (\overrightarrow{h\textsubscript{t}},  \overleftarrow{h\textsubscript{t}}). Hence, all hidden states serve as inputs to the 2 dense fully connected layers, out of which the final layer performs the classification.

\begin{table}[]
\centering
\begin{tabular}{|l|l|}
\hline
\textbf{Parameter}        & \textbf{Value}  \\ \hline
\multicolumn{2}{|c|}{\textbf{Embedding Training}}    \\ \hline
Embedding Size            & 300             \\ \hline
Window Length             & 10              \\ \hline
Sampling Polarity         & Negative        \\ \hline
\multicolumn{2}{|c|}{\textbf{CNN Training}}          \\ \hline
Dropout                   & 0.5             \\ \hline
Stride                    & 1               \\ \hline
\#Kernels                 & 200             \\ \hline
ks\textsubscript{1}                       & 3               \\ \hline
ks\textsubscript{2}                       & 6               \\ \hline
ks\textsubscript{3}                       & 9               \\ \hline
ks\textsubscript{4}                       & 12              \\ \hline
\multicolumn{2}{|c|}{\textbf{RNN Training}}          \\ \hline
\#LSTM Units              & 150             \\ \hline
Input State Dropout       & 0.2             \\ \hline
Recurrent State Dropout   & 0.2             \\ \hline
\multicolumn{2}{|c|}{\textbf{Transformers Fine Tuning}} \\ \hline
Learning Rate             & 1e-5            \\ \hline
Epsilon (Adam optimizer)  & 1e-8            \\ \hline
Maximum Sequence Length   & 256             \\ \hline
Batch Size                & 3               \\ \hline
\#Epochs                  & 5               \\ \hline
\end{tabular}
\caption{Parameter Values}
\label{tab:my-table}
\end{table}

\begin{table*}[]
\centering
\begin{tabular}{|l|c|c|c|c|}
\hline
\multicolumn{1}{|c|}{\multirow{2}{*}{\textbf{DL  Models}}} & \multicolumn{2}{c|}{\textbf{\begin{tabular}[c]{@{}c@{}}Hinglish\\ Data\end{tabular}}}   & \multicolumn{2}{c|}{\textbf{\begin{tabular}[c]{@{}c@{}}Hinglish + English\\ Data\end{tabular}}} \\ \cline{2-5} 
\multicolumn{1}{|c|}{}                                     & \multicolumn{1}{l|}{\textbf{Word2Vec}} & \multicolumn{1}{l|}{\textbf{FastText}} & \multicolumn{1}{l|}{\textbf{Word2Vec}}     & \multicolumn{1}{l|}{\textbf{FastText}}     \\ \hline
CNN                                                        & 62.22                                      & 62.24                                      & 63.00                                          & 62.52                                          \\ \hline
LSTM                                                       & 63.83                                      & 63.69                                      & 64.04                                          & 64.59                                          \\ \hline
Bi-LSTM                                                    & 64.80                                      & 66.02                                      & 65.32                                          & 66.37                                          \\ \hline
Bi-LSTM attention                                          & 66.65                                      & 67.34                                      & 67.44                                          & 68.29                                          \\ \hline
BERT                                                       & \multicolumn{4}{c|}{\textbf{71.43}}                                                                                                                                                       \\ \hline
RoBERTa                                                    & \multicolumn{4}{c|}{70.06}                                                                                                                                                                \\ \hline
ALBERT                                                     & \multicolumn{4}{c|}{66.22}                                                                                                                                                                \\ \hline
\end{tabular}
\caption{Accuracy of Deep Learning Models}
\label{tab:my-table}
\end{table*}

\subsubsection{Transformer Based Models}
{\bf BERT (bert-base-uncased)}: \cite{bert} Being a bidirectional transformer based model pre-trained on a large Wikipedia and Toronto Book Corpus, BERT makes use of a combination of objectives which are meant for the next sentence prediction and masked language modeling tasks.
{\bf RoBERTa (roberta-base)}: \cite{roberta} With some modifications to the parameters of BERT, i.e. changing key hyperparameters, removing the next sentence prediction objective, and training with larger learning rate and batch size values, RoBERTa is built on top of BERT.
{\bf ALBERT (albert-base-v2)}: \cite{albert} Trying to increase the training speed and decrease the memory utilization of BERT, ALBERT is another variation of BERT which repeats layers which are split among groups and splits the embedding matrix into two.

\section{Experimental Settings}

A split of ten percent was made on the total training dataset and the model was trained for a total of 20 epochs. At each epoch, we saved the model checkpoints, and particularly used that checkpoint which was saved before the model begins to overfit to calculate the metrics on the ten percent test dataset split.

Different hyper parameters are involved for the task of training embeddings as well as the models. After working with several optimizers, loss functions and activation functions, the adam optimizer with categorical cross entropy loss function produced the best results for all stated deep learning models. We used relu activation function for all the layers except the output layer, which has sigmoid activation function. We evaluated the performance of CNN models with different values for kernel sizes, activation functions, number of kernels, dropouts, strides and optimizers.

We use pre-trained models like bert-base-uncased, roberta-base, and albert-base-v2, to fine tune the transformer based models on our dataset. Hugging-face\footnote{\url{https://huggingface.co/transformers/}} API was used to fine tune all the transformer based models. Table 2 denotes the hyperparameter combinations used in the training of embeddings, CNN, RNN and fine tuning transformer based models.




\section{Results}

Using all features, \cite{22} show that the baseline model i.e. the SVM classifier with RBF kernel, presented an accuracy of 58.2\%, when dealing with the same emotion labels as we deal with in this paper. 
In the domain of emotion detection in Hindi-English code-mixed data, as far as we know, we are the first to compare transformer based models and word representations. In table 4, the results of CNN, RNN based models for both Word2Vec and FastText based word representations, along with those of transformer based models have been presented. All proposed deep learning models yield better results than state-of-the-art models which deal with these six emotion labels. The best accuracy of 71.43\% is achieved with BERT, as expected. All models utilizing embeddings trained on Hinglish plus English data, perform better than those using embeddings trained on Hinglish data. One conceivable reason for this observation can be the extra coverage of semantics and correlation between the word vectors of English information, which can be utilized for code blended Hinglish information, hence serving as prior data for Hinglish embeddings information. The method works practically equivalent to a knowledge transfer step in which embeddings for English information are utilized as earlier information for embeddings of Hinglish information. Additionally, increased accuracies of all models in case of FastText embeddings, as compared to the Word2Vec embeddings, is observed. One possible reason for this could be the existence of code blended information where FastText enables the coverage of code-mixed vocabulary as against word2Vec which works only on the basis of overall context of word. The transformer based BERT model clearly outperforms both CNN and RNN based models, majorly because of its profound efficiency and its ability to process the input out of order.

The major obstacles in the task of detecting emotions in Hindi-English code-mixed data are handling the linguistic complexities associated with code-mixed data and absence of clean data. Thus, we require even more class-specific cleaner data, in order to reduce the effect of noise, which comes from spelling mistakes, stemming words and the presence of multiple contexts.


\section{Conclusion}

As recent years have seen the rise in usage of social media for open expression of stance and opinions, sentiment analysis and opinion mining have gained attention as problems and become primary areas of research. 

In this paper, we present an openly available class-balanced dataset of Hindi-English code-mixed data, consisting of tweets belonging to 6 types of emotions, which are happiness, sadness, anger, surprise, fear and disgust. We contrast the performance of two types of word representations, both trained on relevant scraped tweets from scratch. We develop two different types of embeddings, one which is trained on solely Hinglish tweets, the other which is trained on a mix of Hinglish and English tweets, and present the performance in both cases. Also, we present deep learning based models including CNNs, RNNs, and transformers, where BERT performs the best among all.

As future scope, the problem can be solved to obtain even better results by carrying out a comparison of MUSE aligned vectors, pre-aligned FastText word embeddings and language specific transformer based word embeddings.

\bibliography{eacl2021}
\bibliographystyle{acl_natbib}

\end{document}